\documentclass{article}
\usepackage{iclr2026_conference,times}
\iclrfinalcopy  

\usepackage[hyphens]{url}
\usepackage{graphicx}
\usepackage{booktabs}
\usepackage{amsmath,amssymb}
\usepackage{xcolor}
\usepackage{enumitem}
\usepackage[colorlinks=true,linkcolor={black!70!blue},citecolor={black!60!green},urlcolor={black!70!blue}]{hyperref}
\usepackage[capitalise,nameinlink]{cleveref}
\crefname{equation}{Eq.}{Eqs.}\Crefname{equation}{Eq.}{Eqs.}
\crefname{appendix}{Appendix}{Appendices}\Crefname{appendix}{Appendix}{Appendices}
\let\cref\Cref

\graphicspath{{figures/}}
\title{Context Is King: How In-Context Specification Shapes the Geometry of Concepts}
\author{%
  Elad David \\
  Zenity \\
  \texttt{eladd@zenity.io} \\
  \And
  Max Fomin \\
  Zenity \\
  \texttt{maxf@zenity.io} \\
}

\begin{document}
\maketitle

\begin{abstract}
Large language models place structured concepts on geometrically faithful
manifolds: weekdays lie on a circle, months on another, usually taken to be a fixed
world-model the network stores and looks up. We show that \emph{context is king}: the
structure a model actually \emph{uses} is set by the in-context specification. A
declarative rule fixes not only \emph{which} relations the geometry encodes but its
\emph{topology type}: the same tokens form a cycle or a branching tree on command,
built even on arbitrary, meaning-free tokens with no prior to inherit, which a
relabeled stored shape cannot do. When the specification conflicts with a strong
pretrained prior, the context-set geometry \emph{dominates} it in capable models, read from the same
activations (representational similarity $0.6$--$0.9$ to the imposed structure versus
near-zero to the prior), across the priors we test and both families we study (Gemma, Qwen).
Activation patching shows the map is \emph{causally used}, not a probe correlate:
swapping one entity's activation for another's makes the model answer with the other
entity's successor under the imposed order. A rough map forms readily, present even in
small and base models; what scale gates is
\emph{using} it cleanly: clean dominance and the causal crossover emerge only in the larger
models (up to Gemma-31B and Qwen-27B) and weaken or reverse below, so a mechanism present in a
large model can be absent in a smaller one of the same family. Whether the model builds this geometry anew or
reconfigures a stored one we leave open; operationally, the geometry it uses is the one
the context specifies.
\end{abstract}

\section{Introduction}
\label{sec:intro}

The geometric structure of concept representations is drawing increasing attention in
interpretability research. The
earliest and still influential answer is the linear representation hypothesis:
a concept corresponds to a direction, and semantic relations to displacements along
such directions \citep{gurnee2023,arditi2024}. It
has since grown from single concepts to how \emph{related} concepts sit together,
which turn out to form geometrically faithful wholes: the days of the week lie on a
circle, with comparable relational geometry for ordered, cyclic, and hierarchical
concepts alike \citep{engels2024,park2024geometry}. Such structure is naturally read
as a stored world-model, geometry the network fixed in pretraining and looks up
whenever a concept is invoked.

\begin{figure}[htbp]
  \centering
  \includegraphics[width=\linewidth]{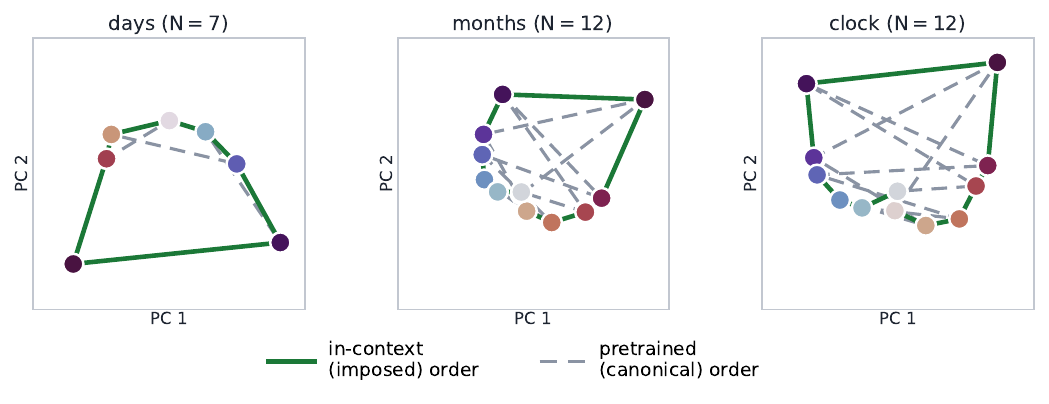}
  \caption{Entity centroids for three cyclic concepts under a conflicting in-context
    order (Gemma-31B; 2D PCA of the last-token pre-generation state; one representative
    scramble, colored by imposed position). Each path is traced in the pretrained
    order (dashed) and the imposed order (solid). Imposed vs.\ pretrained RSA
    (full space, mean over 10 scrambles): days $+0.87/{-}0.03$, months $+0.81/{+}0.05$,
    clock hours $+0.82/{-}0.05$.}
  \label{fig:flip}
\end{figure}

In-context exemplars can reorganize this geometry, though a strong pretrained prior has
been found to resist \citep{park2025}; we show a \emph{declarative} specification decides
the shape outright: the asserted relations, not the prior, set the geometry the model
uses.
We force context and prior into conflict: we declare a redefined order
(``the only valid order is: Wednesday, Monday, \ldots'') and read the
pre-generation representation. The pretrained manifold gives way to the
imposed one (\Cref{fig:flip}); a different relational specification (a
hierarchy rather than an order) yields a different topology altogether
(\Cref{sec:topology}).

This bears on a basic question about how language models represent concepts: is a
concept's relational structure an intrinsic property the model stores and
retrieves, or is it assembled from the context on demand? A prominent line of work reads this structure realistically, as a stored world-model the network looks up \citep{engels2024,gurnee2023,park2024geometry,li2022,nanda2023}, taking the recovered geometry as a fixed fact about the model.
Our results complicate that picture: the geometry a model \emph{uses} is the one
the context specifies, and the stored structure, though real, gives way the moment
the context contradicts it. This is mechanistic, not instruction-following: the
geometry's \emph{type} changes on command, forms on tokens with nothing to retrieve,
and flips under a causal edit. We establish this in two parts: that context sets the
geometry the model \emph{represents} (\Cref{sec:represents}), and that
the context-set map is \emph{causally used}, and its clean form is scale-gated (\Cref{sec:causal}). Our contributions:
\begin{enumerate}[leftmargin=*,itemsep=2pt]
 \item \emph{The specification sets the topology \emph{type}, not merely a shift
 within a fixed shape.} A declarative rule fixes whether a concept's geometry is a
 cycle or a branching tree, and this is the geometry the model \emph{uses}, appearing even under
 prompts that ask nothing structural (\Cref{sec:topology}).
 \item \emph{On conflict, the context-set geometry dominates, and its map is causally
 used.} A declarative rule makes the used geometry follow the context, not the pretrained
 prior (anisotropy and crossover controls, \Cref{sec:redraw}), and an entity-substitution
 patch shows that map is causally operative, not a probe correlate (\Cref{sec:causal}).
 \item \emph{Forming is cheap; faithful use is capability-gated, a transfer caution.}
 Small and base models form a rough imposed geometry, but the clean, traversable map and
 full causal double dissociation emerge only with scale (\Cref{sec:capability}), so a
 small-model finding must be verified at scale before assuming it transfers up.
\end{enumerate}

All code, cached data, and interactive 3D explorers are available at
\url{https://github.com/eladd-ai/context-is-king}.

\section{Background and Related Work}
\label{sec:related}

The question of what sets the relational geometry a model represents over a set of
concepts connects three lines of work: how concepts are encoded geometrically,
how in-context information reshapes those representations, and how such structure is
used mechanistically. We take each in turn and position our claim against it.

\paragraph{Geometry of concepts.} Periodic concepts occupy low-dimensional cyclic
manifolds \citep{engels2024}, many properties are linearly decodable
\citep{gurnee2023}, and categorical/hierarchical concepts form simplices and
polytopes (convex arrangements whose vertices are the categories)
\citep{park2024geometry}. These are
fixed readouts of pretrained structure, whether a concept is a causal direction
\citep{arditi2024} or a manifold whose geometry has been tied causally to behavior
\citep{wurgaft2026}. We instead ask how the \emph{context} sets
relational structure over many entities, testing that reading against a directly
conflicting specification, and the geometry yields to context.

\paragraph{Causally-validated world models.} The strongest
realist evidence is intervention, not probing: in Othello-GPT, linear probes
recover a board state that \emph{causally} steers play when edited
\citep{li2022,nanda2023}. We do not dispute that learned structure is causally
real when uncontested; our own patching (\Cref{sec:causal-use}) confirms a
relational map is used. But those studies probe structure the model was trained
to track and never conflict it with the context, so causal use cannot distinguish
a fixed store from a map constructed by default. Under forced conflict the
causally-effective map follows the \emph{context}: the stored relational structure,
where one exists, is overwritten on demand.

\paragraph{Context-induced geometry.} \citet{park2025} force conflict from the
other side: teaching an in-context graph by random-walk \emph{exemplars} on a semantic
concept (the days ring), they find the pretrained structure resists, holding the top
principal components while the context structure reaches only lower ones. Two things distinguish
our setup, and plausibly account for the shift to dominance: the structure is a
\emph{declarative rule} with no exemplars to learn from, and we read the integrated
pre-generation state rather than the concept cloud's top components. We further set the \emph{topology type} by specification
and confirm the context-set map causally, neither asked in this line. Two concurrent works pursue
context-induced reshaping without forcing the conflict: \citet{hosseini2026} find
naturalistic context \emph{straightens} neural trajectories, and \citet{xiong2026} reorganize geometry along in-context axes. They independently corroborate that context reshapes geometry, but with no
competing prior to override and only correlational links to behavior.

\paragraph{Scale and emergence.} Emergence debates \citep{wei2022,schaeffer2023} turn
on thresholded benchmark scores; our quantities are continuous (imposed-order RSA) and
causal (a patch flip), and their emergence is layered
(forming cheap, clean/traversable use scale-gated; \Cref{sec:capability}). This echoes,
in the representation, the gap \citet{lepori2026} report between encoding a structure and
deploying it, and bears on whether circuit-level interpretability transfers across size
\citep{lieberum2023}.

\paragraph{Mechanistic prior work.} Activation patching was developed to locate
\emph{parametric} facts in the weights \citep{meng2022}; we turn it on structure that
exists only in the context (a scrambled order is nowhere in the weights), shifting the
question from \emph{where} a circuit lives to \emph{whether} the computation forms. The
closest work \citep{kim2025} patches and dissociates across a family sweep but asks
\emph{which} mechanism (schema vs.\ binding), not \emph{when} richness emerges.

\section{Framework}
\label{sec:framework}

We frame the paper around one question: is a concept's relational geometry fixed, or
is it set by the context's specification? This section defines the objects the question
is about and the space of specifications we impose; how we \emph{measure} the geometry
is deferred to \Cref{sec:setup}.

\begin{figure}[htbp]
  \centering
  \includegraphics[width=\linewidth]{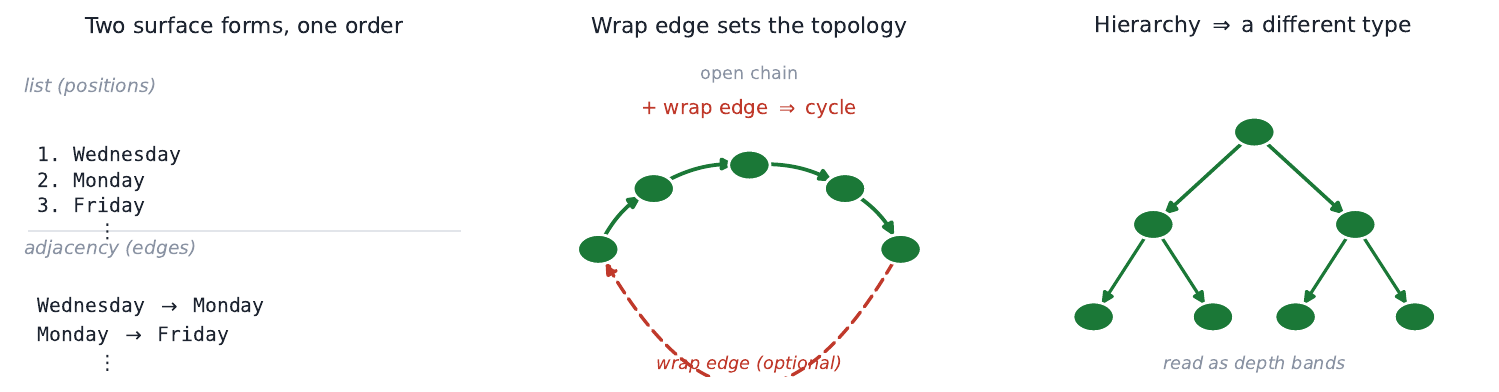}
  \caption{\textbf{What a specification is.} The same order stated as a numbered
    list or as adjacency edges (left); adding a \emph{wrap} edge closes the chain
    into a cycle (middle; the open, no-wrap case is a weaker baseline); a declarative
    \emph{hierarchy} is a different structure type, read as depth layers (right).}
  \label{fig:defs}
\end{figure}

\subsection{Definitions}
\label{sec:definitions}

\textbf{In-context specification.} A declarative rule, given with no exemplars,
asserting the \emph{relations} among a set of entities, an order (``the only valid
order is: Wednesday, Monday, \ldots'') or a hierarchy (``$P$ is the parent of $A$
and $B$''), and, on conflict with the entities' pretrained structure, that the
normal meaning no longer applies.

\textbf{Concept geometry.} For an entity set $E=\{e_1,\dots,e_N\}$, we represent each
entity by a centroid $c_i$ (its mean representation) and describe the geometry by the
pairwise-dissimilarity matrix $D_{ij}=d(c_i,c_j)$. Its \emph{topology} is the shape of
that arrangement, an open line, a closed cycle, or the depth layers of a tree.

\textbf{The geometry the model \emph{uses}.} We read entity representations from the
residual stream at the last prompt token, immediately before generation, the integrated
state the next-token computation consumes, not one conjured by a structural query or tied
to a single paraphrase. A geometry is \emph{used} if the model's behavior and causal
mechanics run on it (\Cref{sec:setup}). The readout is a substantive choice: the token-local state can still carry the
pretrained order (\Cref{fig:possweep}), and activation patching (\Cref{sec:causal-use})
confirms the integrated state is the one the model acts on.

\textbf{Congruent vs.\ conflicting specification.} A specification is
\emph{congruent} when its relations agree with the pretrained structure (the natural
weekday order) and \emph{conflicting} when they contradict it (a scrambled, redefined
order). Conflict is our most direct test, forcing a choice between the stored
structure and the stated one.

\subsection{Specification taxonomy}
\label{sec:taxonomy}

Every experiment is one point in a common space: impose structure $X$ on entities
$Y$, congruent or conflicting, in surface form $Z$:
\begin{itemize}[leftmargin=1.4em, itemsep=1pt, topsep=2pt]
 \item \textbf{Structure type:} an \emph{order} (open line, or closed cycle once a
 wrap relation is stated) or a \emph{hierarchy} (binary tree).
 \item \textbf{Relation to the prior:} \emph{congruent}, or \emph{conflicting} (a
 \emph{scramble}, a fresh random permutation of the order, or a redefinition).
 \item \textbf{Entities:} a gradient of pretrained prior strength, \emph{strong}
 cyclic orders (weekdays, months, clock hours), a \emph{weak} cyclic prior (zodiac)
 and a \emph{non-cyclic} one (musical notes, linear by pitch), and \emph{arbitrary
 structure-free tokens} (strength measured as no-context ring RSA, \Cref{sec:redraw}).
 \item \textbf{Surface form:} a numbered \textbf{list} or \textbf{adjacency} edges,
 with or without the \textbf{wrap} edge that closes a loop; form $\times$ wrap (a
 2$\times$2) separates \emph{structure} from \emph{surface form} (\Cref{fig:defs}).
\end{itemize}
The central case is a \emph{conflicting order} on a real cyclic concept; the
congruent, hierarchical, and arbitrary-token cases map out the rest
(\Cref{sec:topology}).

\section{Experimental Setup}
\label{sec:setup}

The preceding section set up the objects; here we fix how we \emph{measure}
the geometry: the probe, the prompt regimes, the models, and the sampling. Its causal
counterpart, activation patching, is presented with its results in
\Cref{sec:causal-use}.

\paragraph{Entity-layout probe.} RSA is standard in computational neuroscience for
comparing representational geometries \citep{kriegeskorte2008}, though uncommon in
mechanistic interpretability. We form one centroid $c_i$ per entity from the
last-token pre-generation residual at a single late layer, averaged over neutral query
templates (questions, commands, narratives; \Cref{app:prompts}), so the geometry is not
a per-phrasing artifact. Some templates ask for an entity's $k$-step successor under the
stated order (e.g., ``1 step after Monday''); others merely mention it. With $\tilde c_i=c_i-\bar c$ the
mean-centered centroid, the representational dissimilarity matrix (RDM) has entries
$D_{ij}=1-\cos(\tilde c_i,\tilde c_j)$, and RSA is the Spearman correlation between
its off-diagonal entries and a theory-specified template $T$ (for a cycle,
$T_{ij}=\min(|i{-}j|,\,N{-}|i{-}j|)$):
\[
 \mathrm{RSA}=\rho_{\mathrm{Spearman}}\!\big(\{D_{ij}\}_{i<j},\,\{T_{ij}\}_{i<j}\big).
\]
Mean-centering is essential: the residual stream is anisotropic, and subtracting the
shared direction is what exposes the ordering (\Cref{app:controls}). The specific layer,
$L=\mathrm{round}(0.75\,n_{\text{layers}})$, is chosen from a full-layer sweep that shows
a broad late-network plateau, so the result does not hinge on the exact choice
(\Cref{app:layersweep}).

\paragraph{Why RSA, and why it behaves in high dimension.} Concept-geometry work
usually reads structure off PCA plots or fitted probes and SAEs
\citep{engels2024,gurnee2023}; we instead score the centroid RDM against a
\emph{theory-specified} template with a permutation null and an
imposed-versus-natural dominance test, the testable form of what such work shows by
eye. Cosine RSA is well-behaved here: with at most twelve centroids the geometry spans
$\le N{-}1$ dimensions (not a high-dimensional cloud), RSA is rank-based, and
mean-centering removes the anisotropic direction (null on the same points; cleared 10/10 scrambles in both large
models). Four independent measures (a circular probe, ring planarity, persistent
homology, behavior) converge on the same structure ($\rho{=}0.93,\,0.83$;
\Cref{app:convergent}).

\paragraph{Regimes.} We read geometry from the state that generates the behavior, the
residual stream at the last prompt token, before generation, under three prompt regimes
that share the rule and query and differ only in the trailing instruction
(\Cref{app:prompts}):
\begin{itemize}[leftmargin=1.4em, itemsep=1pt, topsep=2pt]
  \item \emph{direct}: answer immediately;
  \item \emph{scripted reasoning}: we dictate a step-by-step trace and read along it (the imposed structure is present at every step; we draw no conclusion from the dictated dynamics);
  \item \emph{free-form}: the model's own scaffold, read at the pre-thinking token.
\end{itemize}
The imposed-order structure forms in all three (\Cref{sec:redraw}). We read free-form
only at the pre-thinking token, not along the model's reasoning trace.

\paragraph{Models.} We study eight instruction-tuned models across three families and
roughly an order of magnitude in scale: Gemma-4 (E2B, E4B, 12B, 31B), Qwen-3.5 (4B,
9B, 27B), and Llama-3.1-8B as a cross-architecture capability floor. The scaling claim
spans all eight; but because a clean, usable map forms only with scale
(\Cref{sec:capability}), our main analyses (causal patching, robustness controls)
concentrate on the two large models that reliably build it, Gemma-31B and Qwen-27B.
For the causal analysis we additionally sweep the full Gemma ladder (2B--31B) to locate
where the effect emerges with scale. Throughout, a paired value $a/b$ is
Gemma-31B\,/\,Qwen-27B.

\paragraph{Sample sizes and intervals.} Geometry RSA is averaged over 10 scrambles per
concept, the imposed hierarchy over 6, each with per-scramble 95\% confidence intervals.
For the causal arm we report per-scramble values: the causal patching test
(\Cref{sec:causal-use}) uses 2 scrambles $\times$ 42 operand pairs per model, in both
direct and chain-of-thought prompting.

\section{Context Sets the Geometry the Model Uses}
\label{sec:represents}

Our representational evidence is twofold: on a conflicting specification the geometry
the model uses follows the stated relations rather than the pretrained prior
(\Cref{sec:redraw}), and the specification fixes not merely \emph{which order} the
entities take but the \emph{topology type} of their arrangement, a cycle or a tree
(\Cref{sec:topology}).

\subsection{The Context's Geometry Wins on Conflict, and Forms Without a Prior}
\label{sec:redraw}

\begin{figure}[htbp]
 \centering
 \includegraphics[width=0.88\linewidth]{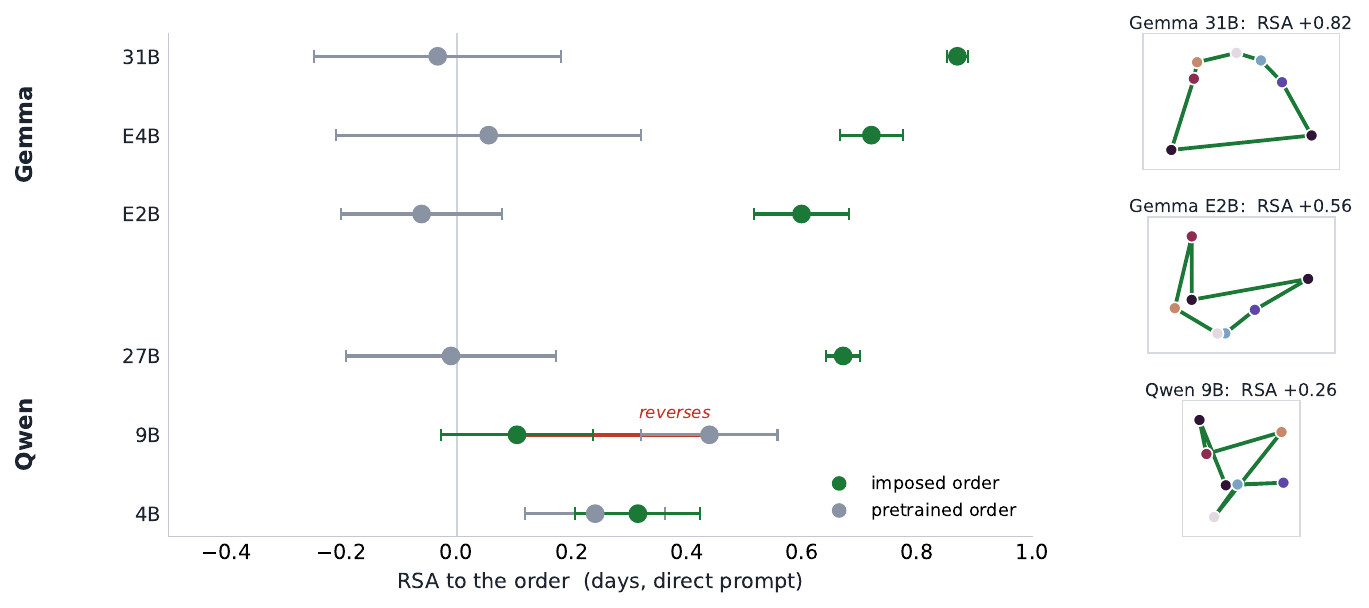}
 \caption{Imposed-order RSA (green) vs.\ residual natural-order RSA (gray) per model
 in the same probed state (\emph{left}; days, direct prompt, per-scramble 95\% CIs),
 and imposed-order geometry for three models (\emph{right}; annotated RSA). Per-model
 values in \Cref{app:redraw}.}
 \label{fig:dominance}
\end{figure}

When the specification is rewritten, the entity-layout geometry reorganizes
to the imposed order while the pretrained one collapses. In the \emph{same} probed
state the imposed order dominates the natural one, for Gemma-31B on days by a gap
of $+0.90$ under a direct prompt and $+0.58$ free-form, with the pretrained ring
suppressed to near zero (\Cref{fig:dominance}; per-model values in \Cref{tab:rsa}).
Imposed-order RSA is high across all five concepts, regardless of their
pretrained-prior strength, and weaker in Qwen than Gemma. The effect is
capability-gated: strong in capable models, it collapses or reverses in the smaller
ones (Qwen-9B days $-0.33$, the natural ring retained; Llama-8B does not build the map
at all), so dominance on conflict is a capable-model claim, not a universal one.

The overridden prior itself varies widely: with \emph{no} specification the natural
ring is strong for weekdays/months/clock ($\approx0.6$--$0.8$), weak for zodiac
($\approx0.3$), and negative for musical notes (a linear pitch scale). Context imposes
its structure across this whole range, and on arbitrary meaning-free tokens builds it
from nothing (\Cref{tab:arb}); whether the induced shape is a ring or a tree is set by
the specification (\Cref{sec:topology}).

\begin{figure}[htbp]
 \centering
 \includegraphics[width=\linewidth]{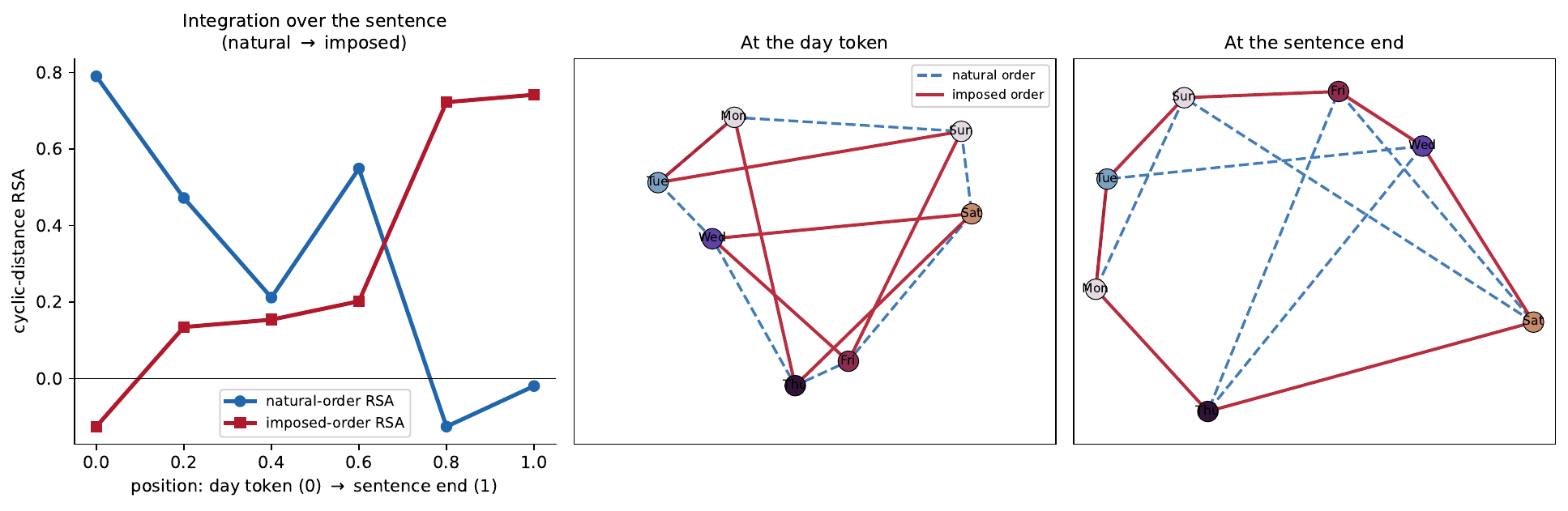}
 \caption{Imposed- and natural-order RSA by readout position (\emph{left}; day token
 $=0$ to sentence end $=1$, Gemma-31B), and the entity centroids at each position
 traced in both orders (\emph{middle, right}; natural dashed, imposed solid).}
 \label{fig:possweep}
\end{figure}

\paragraph{Artifact controls.} The dominance survives the
checks that could show it inflated by the metric or summoned by our question (full
procedures in \Cref{app:redraw,app:controls,app:cooccur}):
\begin{itemize}[leftmargin=1.4em, itemsep=3pt, topsep=3pt]
 \item \textbf{Relabeling.} When the imposed order
 changes the geometry moves to match it (a crossover a fixed shape cannot produce), and
 among all $5040$ orderings of the seven days the geometry fits the \emph{imposed} one
 best, not the natural ring.
 \item \textbf{Anisotropy.} Against a cone-preserving null (\Cref{app:controls}) all ten
 scrambles clear on every one of the five concepts in both large models, the
 twelve-point concepts at $p<0.001$, and the smallest, seven-point concepts
 (days, notes) at $p\le0.008$, with natural-order retention near zero, so the
 residual-stream cone does not explain the effect.
 \item \textbf{Stated list position.} The geometry does not follow list position: the
 imposed tree (\Cref{sec:topology}) never states depth yet its depth structure appears
 and survives shuffling siblings apart, and in the imposed cycle the first and last
 items sit adjacent though they are farthest apart in the list.
 \item \textbf{The probe itself.} The ordering is not summoned by our question. It
 appears even when the prompt asks nothing structural (a bare mention or a short
 story), where the imposed order still dominates the natural one (\Cref{tab:robust}); the same geometry emerges whether the query asks for one step or the
 full traversal (\Cref{tab:robust}, \Cref{fig:kmarg}); and it forms only after the model
 reads the redefinition, handing off from the day token (still the pretrained order) to
 the sentence end (the imposed order; \Cref{fig:possweep}).
\end{itemize}

\subsection{The Specification Sets the Topology Type}
\label{sec:topology}

\Cref{sec:redraw} showed context fixes \emph{which order} the geometry encodes;
here it fixes something stronger: the \emph{topology type}. A model that
relabeled a stored ring could reorder it but never turn it into a tree;
changing only the asserted relations does exactly that, ruling out lookup of a
\emph{fixed} stored structure. The clearest case is a hierarchy (a tree, which no
reordering of a ring could produce), so we take it first; we then hold the tokens fixed
and vary the declared relation (cycle versus tree), and ask how firmly the stated wrap
closes the ring.

\begin{figure}[htbp]
  \centering
  \includegraphics[width=\linewidth]{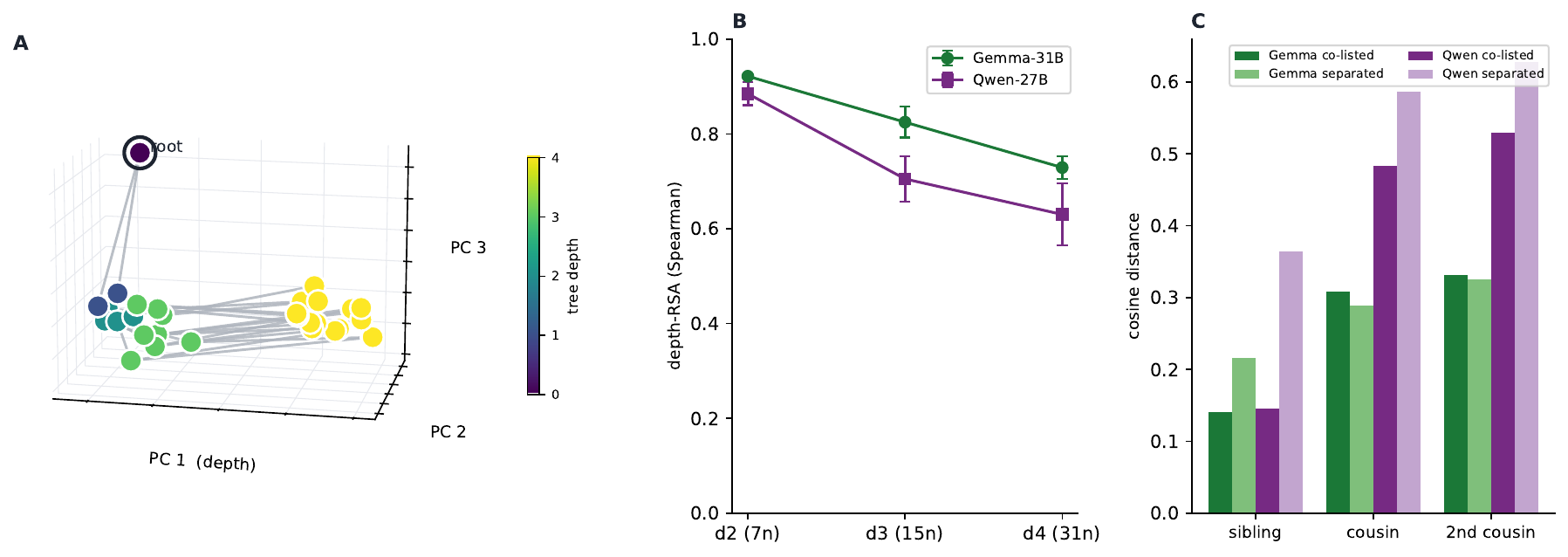}
  \caption{Imposed depth-4 tree on arbitrary tokens under neutral queries
    (Gemma/Qwen). \emph{(A)} Entity-centroid PCA (color $=$ depth; parent/child edges;
    depth-RSA $+0.73$). \emph{(B)} Depth-RSA as the tree grows, $+0.92/+0.89$
    (depth-2) to $+0.73/+0.63$ (depth-4; per-scramble 95\% CIs). \emph{(C)} Among
    same-depth leaves, cosine distance vs.\ common-ancestor distance; co-occurrence
    control depth-RSA $0.83\to0.82$ (Gemma), $0.71\to0.72$ (Qwen).}
  \label{fig:hierarchy}
\end{figure}

\paragraph{A tree on arbitrary tokens.} A declarative binary tree on arbitrary tokens
forms a depth-stratified manifold: under neutral queries the centroid RDM tracks
node depth with RSA near $+0.9$ at depth-2, holding to $+0.73/+0.63$ (Gemma/Qwen)
at depth-4 (\Cref{fig:hierarchy}A,B). Same-depth nodes band together, parent and
child are the farthest pairs.

The finer branch relation is real but only metric (\Cref{fig:hierarchy}C): among
same-depth leaves, cosine distance grows with common-ancestor distance; we do not claim
a faithful tree embedding. A co-occurrence control (siblings shuffled apart) leaves
depth-RSA unchanged but removes part of the raw sibling tightness, so the relation is
partly surface-driven (\Cref{app:prompts}).

The hierarchy is constructed, not an artifact of probing or word similarity: it is
\emph{intrinsic} (strengthening as the query asks less, peaking near $+0.9$ when
nothing structural is asked), \emph{not token similarity} (under re-randomized token$\to$node
assignments, a pair is closer when placed as siblings than when not, 12/12), \emph{capability-gated} (the
2B model builds none), and \emph{query-invariant} (across 36 query types the tree
relocates bodily, the query accounting for 97\% of positional variance, while the
per-node consensus shape is preserved, $\rho{=}0.70$; \Cref{fig:queryreloc} in
\Cref{app:redraw}).

\begin{figure}[htbp]
  \centering
  \includegraphics[width=0.88\linewidth]{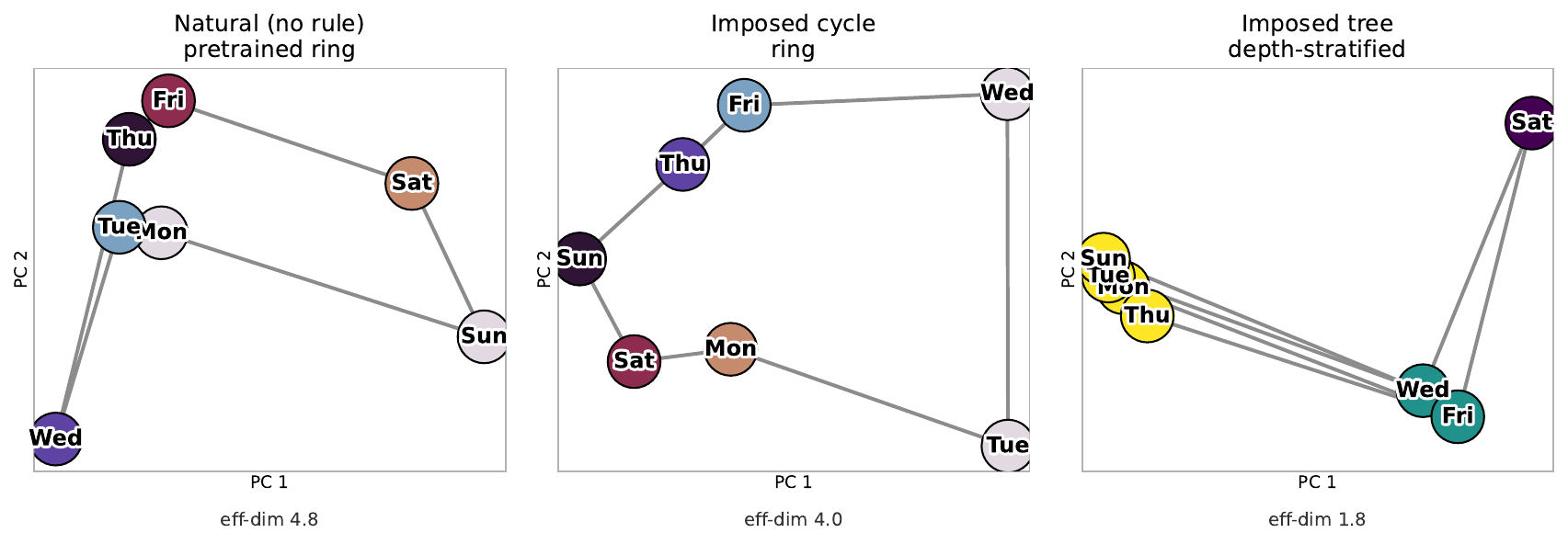}
  \caption{Seven weekdays under three specifications (Gemma-31B; per-condition 2D
    PCA, one scramble; effective dimension $=$ participation ratio, mean$\pm$std over
    10 scrambles): un-imposed (day-token readout), an imposed cycle ($\approx$4-D),
    and an imposed tree ($\approx$1.8-D). Cross-family values and the line case in
    \Cref{app:shapes}.}
  \label{fig:shapes}
\end{figure}

\paragraph{Same entities, different shape on command.} We hold the
\emph{entities} fixed and vary only the declared relation. On the seven weekdays (un-imposed, a pretrained ring) an imposed cycle keeps a ring (imposed-order
RSA $+0.72/+0.71$, effective dimension $\approx$4) while an imposed \emph{tree}
restratifies the \emph{same} days by depth (depth-RSA $+0.67/+0.68$, cyclic
structure gone, effective dimension $\approx$2); both read at the integrated
sentence-end over 10 scrambles, across both families (\Cref{fig:shapes}). The tokens
never change, so it is the specification, not the concept, that sets the topology
\emph{type}, a distinction that also shows up in a relabeling- and rotation-invariant
persistent-homology test (a dominant $H_1$ loop for the imposed cycle, none for the
tree; \Cref{app:convergent}), not only in the rank metric. An imposed \emph{line} opens
the ring only partially and capability-dependently (\Cref{app:shapes}).

\paragraph{The wrap relation strengthens closure.} Within cyclic concepts, stating
the \emph{wrap} relation (the last item returns to the first) strengthens closure into
a ring, a graded, directional effect (pronounced in Gemma, weaker in Qwen), driven by
the asserted relation rather than surface format or endpoint co-occurrence
(form~$\times$~wrap 2$\times$2, \Cref{app:cooccur}).

\section{The Context-Set Map Is Causally Used, and Its Clean Form Is Scale-Gated}
\label{sec:causal}

\Cref{sec:represents} showed the context sets the geometry the model
\emph{represents}; but a represented structure could be epiphenomenal. We now show the
\emph{same} map is \emph{causally} operative (\Cref{sec:causal-use}), characterize the
scale it takes to use it \emph{cleanly} (\Cref{sec:capability}), and localize where it
is built to a mid-network identity$\rightarrow$binding stage (\Cref{app:localize}).

\subsection{Activation Patching Steers the Answer Through the Constructed Map}
\label{sec:causal-use}

\begin{figure}[htbp]
  \centering
  \includegraphics[width=\linewidth]{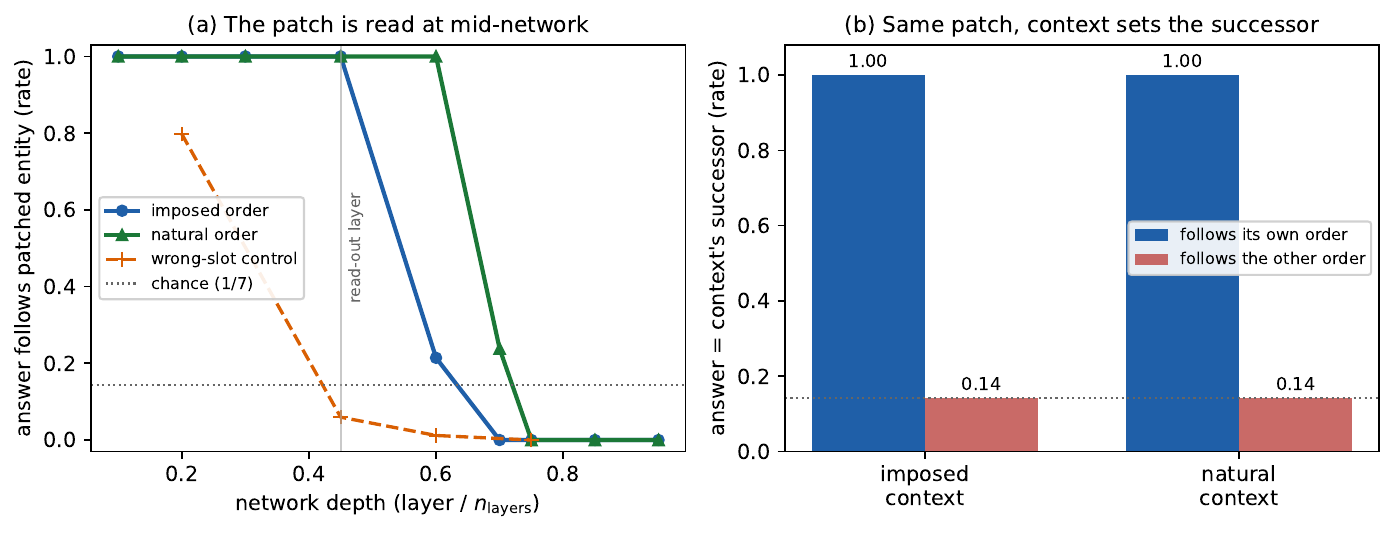}
  \caption{\textbf{The context-set map is causally used} (Gemma-31B, days; entity-substitution
  patch $h_{E_{\text{orig}}}\!\leftarrow\!h_{E_{\text{patch}}}$, 42 ordered pairs $\times$ 2
  scrambles). \textbf{(a)}~Across depth, the patched answer follows the donor entity's
  successor when the overwrite is applied at mid-network, under both the imposed and the
  natural order; the wrong-slot control (writing the entity vector at a non-entity slot)
  leaks only at the earliest layers and then falls to chance, locating the effect at the
  entity. \textbf{(b)}~At the read-out layer the \emph{same} patch follows whichever
  successor the \emph{context} specifies---imposed-order under the imposed context,
  pretrained-order under the natural context---and the other order only at chance
  ($1/N\approx0.14$): a double dissociation. The answer is computed from the (patched)
  representation through the context-built map, not read off the surface token.}
  \label{fig:crossover}
\end{figure}

We test the map by \emph{intervention}. At one layer we overwrite the \emph{orig} entity's residual (the entity the prompt
asks about) with the \emph{patch} entity's activation, a donor from the same context:
$h_{E_{\text{orig}}}\!\leftarrow\!h_{E_{\text{patch}}}$. We read the answer and, over 42
ordered pairs, score the fraction equal to $\mathrm{succ}(E_{\text{patch}})$, the patch's
successor, under the
imposed and under the pretrained order (chance $1/N\approx0.14$ for seven days). The
result is a clean \emph{crossover} (Gemma-31B; \Cref{fig:crossover}): imposed context $\to$
imposed-order successor ($1.00$); natural context, \emph{same} patch $\to$
pretrained-order successor ($1.00$); each other order only at chance. The answer
follows whichever map the context sets, so the entity is causally read, not a passive
correlate: its identity is carried forward and the context's ordering is the transform
applied to it. We characterize this patch across both the Gemma (E2B--31B) and Qwen (4B--27B) ladders,
in direct and chain-of-thought prompting (\Cref{tab:causalladder}),
and find its strength scales: the \emph{clean} double dissociation (both arms at ceiling
at a deep, stable locus) is reached only by the larger models (Gemma-12B/31B, Qwen-27B),
while at 2--9B the patch is effective only at early layers and is one-sided or shallow. In the larger models the
step count is likewise causally used and an injected operand propagates into a
chain-of-thought trace; the patch also commits deeper with scale ($0.46\rightarrow0.70$
of network depth; \Cref{app:patching}).

The crossover rules out a stored lookup: a fixed table would answer identically
regardless of context, but ours flips with the context, so the map is a
\textbf{computation, not a lookup}, built on demand (and by the topology result, a store
could not change its own shape in any case). It holds under chain-of-thought too: a
verbose reasoner (Qwen-27B) still commits to the imposed successor ($1.00$; pretrained
$\le0.14$; LLM-judged, \Cref{app:patching}). The patch is also a \textbf{localization}:
it pins what is causal (the entity's identity, read through the imposed relation) and
its depth, but not which geometric direction carries it (the manifold axis;
\Cref{sec:limitations}).

\subsection{Forming Is Cheap; Faithful Use Is Scale-Gated}
\label{sec:capability}

What scale gates is not \emph{forming} a rough map, nor reading it \emph{roughly}, but
using it \emph{cleanly and faithfully}. Even a 2B base model clears the null on the
twelve-point concepts (\Cref{tab:rsa}), non-RLHF base models reorganize
on conflict (Gemma-31B base $+0.62$ months, $+0.67$ zodiac, so it is not an
instruction-tuning artifact), and the imposed map is causally readable at small scale,
though only at early layers (\Cref{sec:causal-use}). What scales is \emph{fidelity}: small
models close only a partial, self-crossing ring, and the patch flips the answer under
only one of the two orders; large, instruction-tuned models close clean circles and flip
it under both: the full crossover of \Cref{sec:causal-use}.
Imposed-order RSA rises with scale within each family (\Cref{tab:rsa}); the smallest
models fail outright (Llama-8B $0.08$) or \emph{reverse} (Qwen-9B), and instruction-tuning
lowers the threshold without being its source.

\paragraph{Behavioral validation.} The models \emph{act} on the redefinition, not only
represent it: capable models follow the imposed order one-shot ($k{=}1=1.00$ for
Gemma-12B/31B and Qwen-27B) while less capable ones revert. One-shot reading recovers
only adjacency; for $k\ge2$ accuracy falls while step-by-step reasoning succeeds, so
the structure is \emph{built} pre-generation but \emph{traversed} a hop at a time
(\Cref{app:behavior}, \Cref{app:regimes}).

\section{Limitations}
\label{sec:limitations}

Our causal evidence is a \emph{localization}: entity-substitution patching identifies
the context-defined operator and its depth, not the manifold axis as the causal carrier
(a linear displacement cannot traverse a cycle; nonlinear steering \citep{wurgaft2026}
is out of scope). The representational geometry is a small object (at most twelve
centroids, one late layer, the pre-generation token), and RSA captures rank-agreement
rather than metric shape, so metric-shape claims rest on the corroborating measures of
\Cref{app:convergent}.

Coverage is set by our model range: the effect scales within both families across the
eight-model ladder, with the causal signature characterized on the Gemma and Qwen
ladders (\Cref{tab:causalladder}), an existence proof over these models rather than a
universal law. A manifold-axis steer, richer relations (partial orders and DAGs), and
the geometry during native reasoning remain for future work.

\section{Conclusion}
\label{sec:conclusion}

In two model families, the relational geometry a capable LLM uses over a set of concepts
is set by the context's specification, not fixed in the weights. A conflicting rule pulls
the geometry off the pretrained prior onto the imposed order; the asserted relations set
its topology \emph{type}, a cycle or a tree; and the resulting map is causally used, its
clean form emerging only with scale. The effect spans entities from a strong pretrained
order (days) to arbitrary tokens with none. How the map arises we resolve only partway:
on arbitrary tokens there is nothing stored to retrieve, so it must be \emph{built} from
the specification, while where a strong prior exists, building it and \emph{reconfiguring}
the stored shape into the imposed one are indistinguishable in our readout, and we leave
that open. The reframing holds regardless: the geometry a capable model uses is not a
fixed property of the model but a computation set by the context, and a recovered
``world-model'' geometry should be read as context-dependent, not intrinsic.


\bibliographystyle{iclr2026_conference}
\bibliography{refs}

\clearpage
\appendix
\crefalias{section}{appendix}\crefalias{subsection}{appendix}

\section{Prompt templates}
\label{app:prompts}
Each prompt concatenates a \emph{rule} (the redefined order or hierarchy), a
\emph{query}, and a regime-specific instruction (\Cref{sec:setup}). Verbatim
templates follow; placeholders are in \textit{italics} and wrapped in square brackets.

\paragraph{Conflicting-order rule (cyclic concepts).} The \emph{list} form:
\begin{quote}\ttfamily\small
You are operating in a REDEFINED calendar. The ONLY valid order is below; the
normal order does NOT apply.\\
Order:\\
1. [\textit{entity at position 0}]\\
2. [\textit{entity at position 1}] \;\ldots\\
Full cycle (wraps): [\textit{order repeated}], joined by ``$\to$''\\
Convention: start = position 0; `k steps after X' = item at position k of the
walk.
\end{quote}
The same order in the \emph{adjacency} form states one edge per relation:
\begin{quote}\ttfamily\small
[\textit{entity 0}] is immediately followed by [\textit{entity 1}]\\
{}[\textit{entity 1}] is immediately followed by [\textit{entity 2}] \;\ldots\\
(wrap) [\textit{last entity}] is immediately followed by [\textit{entity 0}]
\end{quote}
List vs.\ adjacency, each with or without the wrap edge (last $\to$ first), are the
factors of the 2$\times$2 (\Cref{sec:topology}).

\paragraph{Query paraphrases (cyclic).} The entity-layout probe marginalizes the
query over the six paraphrases in \Cref{tab:queries} \emph{and} over the step
count $k \in \{1,\dots,6\}$ (36 queries per entity). The one-shot accuracy
analyses (\Cref{app:localize}) use the $k{=}1$ case. These are the \emph{structural}
step-queries; beyond them the probe also reads \emph{non-structural} templates that ask
nothing about the order (a bare mention or a short story; \Cref{tab:robust}).

\begin{table}[htbp]
\centering\small
\caption{Query paraphrases marginalized over by the entity-layout probe (entity
\textit{e}, step count \textit{k}); the probe averages over these and over
$k \in \{1,\dots,6\}$.}
\label{tab:queries}
\begin{tabular}{@{}r l@{}}
\toprule
 & Query (parametric in \textit{k})\\
\midrule
1 & What is \textit{k} steps after \textit{e}?\\
2 & \textit{k} steps after \textit{e} is?\\
3 & From \textit{e}, advance \textit{k} steps. Which item?\\
4 & Starting at \textit{e}, move \textit{k} steps forward. Result?\\
5 & \textit{e} plus \textit{k} steps =?\\
6 & Advance \textit{k} from \textit{e}. Which one?\\
\bottomrule
\end{tabular}
\end{table}

\paragraph{Imposed-hierarchy rule (tree).}
\begin{quote}\ttfamily\small
You are operating under a REDEFINED hierarchy. The ONLY valid parent/child
relations apply below; the normal meanings of these words do NOT apply.\\
Hierarchy:\\
- \textit{P} is the parent of \textit{A} and \textit{B}. \;\ldots\\
Convention: the root has no parent; `steps up to the root' = number of
parent-links from the item to the root.
\end{quote}
The co-occurrence control (\Cref{sec:topology}) instead states one edge per line
(``\textit{P} is the parent of \textit{A}.'') and shuffles them so siblings are
not co-listed. Tree queries are \emph{relational} (``Who is the parent of
\textit{e}?''; ``How many steps from \textit{e} up to the root?'') or
\emph{neutral} (``Consider the item \textit{e}.''; ``Take note of the item
\textit{e}.''), the latter asking nothing structural.

\paragraph{Arbitrary tokens.} The structure-free conditions assign the imposed
relation to everyday English nouns with no inherent cyclic or hierarchical order
(Apple, River, Chair, Cloud, Tiger, Bridge, Lamp, \ldots, extended to twelve for
the larger cycle and to thirty-one for the depth-4 tree). The nouns are chosen so
that no two share subword pieces that could induce morphological similarity, and
they carry no pretrained relation for the model to retrieve, so any recovered
topology must be built from the specification rather than read off the tokens.
For the arbitrary-token conditions, the three regimes (\Cref{sec:setup}) are realized
by these verbatim trailing instructions:
\begin{itemize}[leftmargin=1.4em, itemsep=1pt, topsep=2pt]
  \item \emph{direct}: ``Answer with ONLY the name, nothing else.''
  \item \emph{scripted reasoning}: ``Work through it ONE STEP AT A TIME along the
    order above, then end with a line exactly: FINAL: \textit{name}''.
  \item \emph{free-form}: no trailing instruction; native thinking enabled.
\end{itemize}

\section{Layer-sweep robustness}
\label{app:layersweep}
\begin{figure}[htbp]
 \centering
 \includegraphics[width=\linewidth]{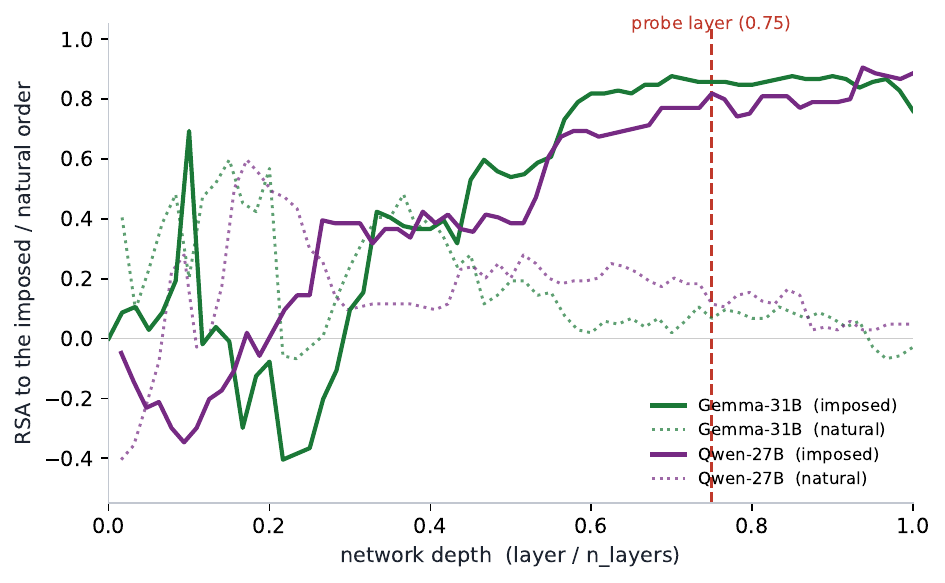}
 \caption{Full-layer sweep (days): imposed-order RSA (solid) and residual
 natural-order RSA (dotted) vs.\ network depth, both flagships; the probe layer
 ($0.75$) marked dashed.}
 \label{fig:layersweep}
\end{figure}

The probe layer is fixed at $0.75\,n_{\text{layers}}$ for every model and
concept; we do not tune it. \Cref{fig:layersweep} reports a full all-layer sweep
of the entity-layout probe (days) for the two flagship models. The imposed-order
RSA is near zero or noisy in the early and middle network and then rises to a
broad plateau across the upper layers, while the natural-order RSA falls to zero
over the same range. The probe layer sits inside this plateau and below each
model's own peak (Gemma-31B 0.86 at $0.75$ vs.\ a 0.88 peak; Qwen-27B 0.80 vs.\
0.91), so the choice is conservative rather than peak-seeking and the redraw
result does not depend on the exact layer.

\begin{figure}[htbp]
  \centering
  \includegraphics[width=\linewidth]{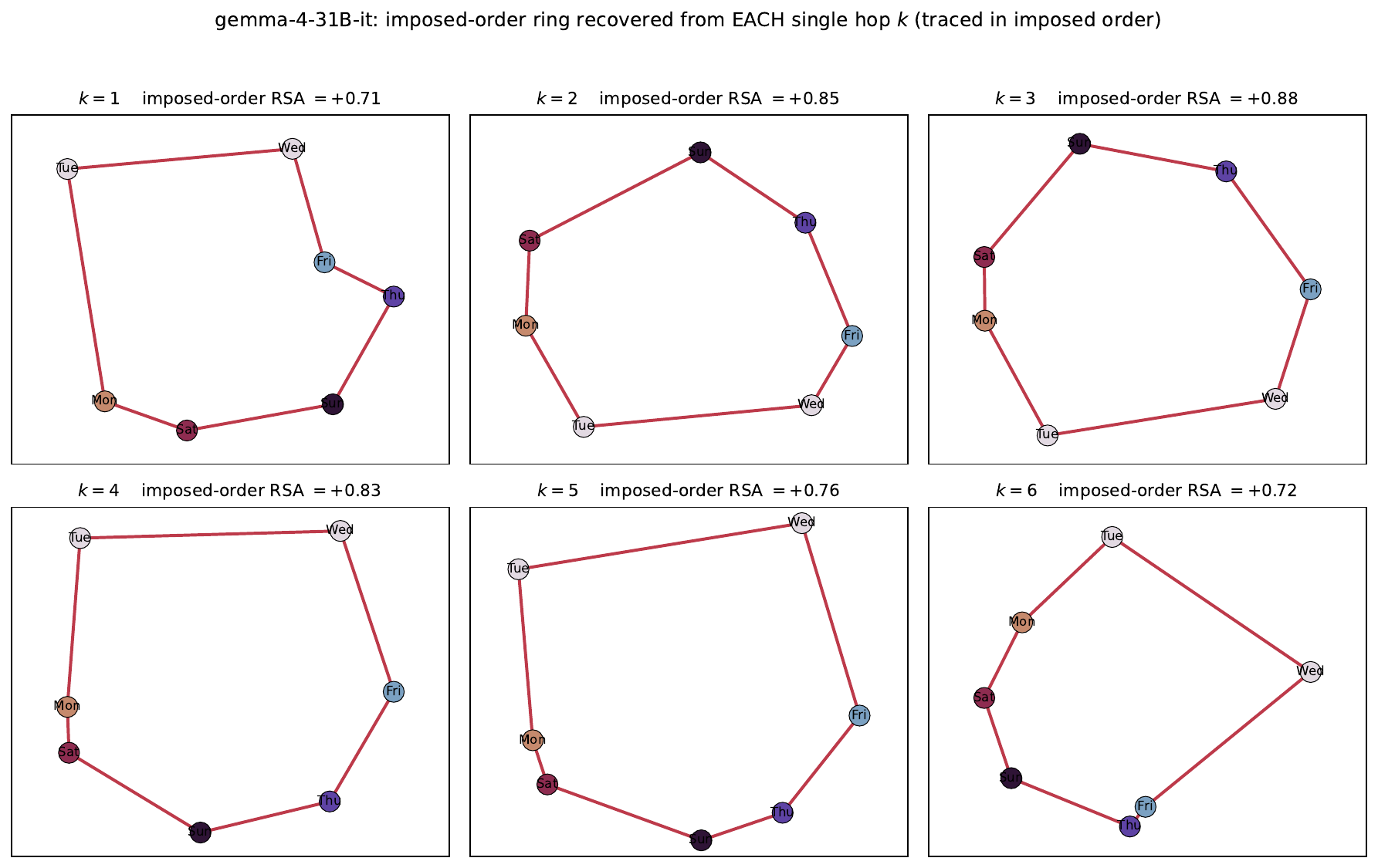}
  \caption{Imposed ring recovered from each hop value $k$ separately (Gemma-31B;
    centroids from one $k$ at a time, traced in the imposed order; imposed-order RSA
    $+0.71$ to $+0.88$ across $k$).}
  \label{fig:kmarg}
\end{figure}

\section{Behavioral validation across the ladder}
\label{app:behavior}
\begin{figure}[htbp]
 \centering
 \includegraphics[width=\linewidth]{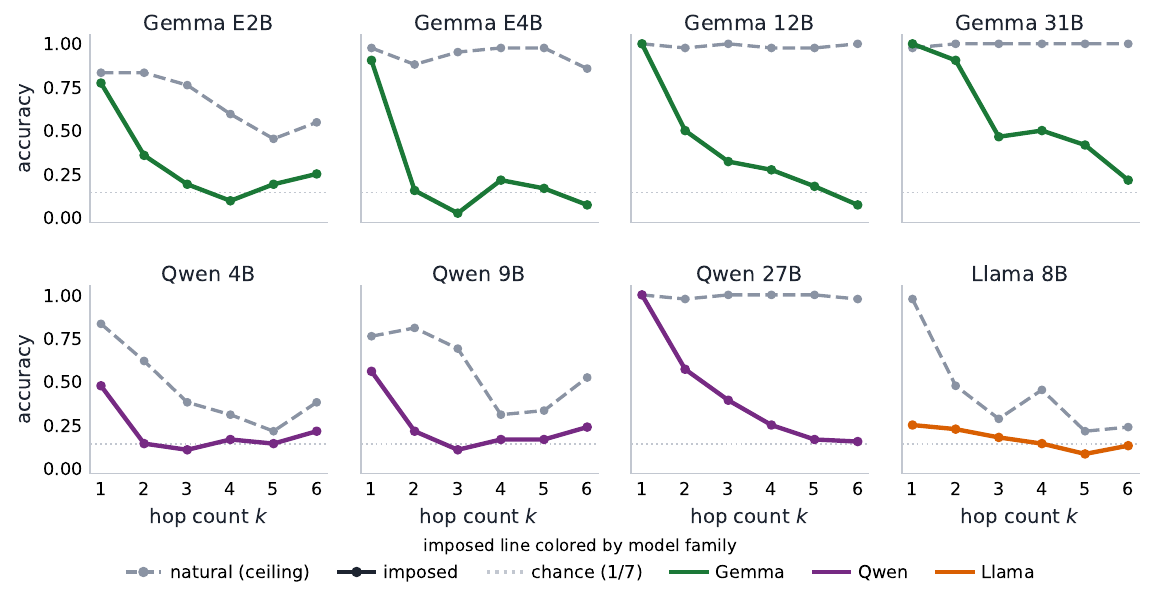}
 \caption{Accuracy by hop count $k{=}1$--$6$ per model in the \emph{direct}
 (no-reasoning) regime (days, list, bare $\le$3-token answers, two scrambles):
 natural order (dashed, per-$k$ ceiling) vs.\ imposed order (solid), with the $1/7$
 chance line. One-shot accuracy degrades with $k$ even for the capable models; the
 reasoning regimes recover ceiling (\Cref{fig:regimes}).}
 \label{fig:behavior}
\end{figure}

\Cref{fig:behavior} plots, per model and at matched hop counts $k{=}1$ to $6$,
accuracy on the natural order and on the imposed order (direct regime; days, list format,
bare $\le$3-token answers, two scrambles). Plotting the natural order at every
$k$, not just the degenerate $k{=}1$, is what makes the comparison meaningful.
For the capable models the natural order stays near ceiling across $k$
(Gemma-12B/31B, Qwen-27B), so their sharp imposed-order falloff at $k\ge2$ is
override-traversal difficulty rather than an inability to traverse at all; for
the weakest models (Llama-8B, Qwen-4B) the natural order itself decays with $k$,
so their imposed failure is partly a base-traversal limit. The geometry, not
these accuracies, is the primary measure.

\section{Behavioral regimes and the generation budget}
\label{app:regimes}
\begin{figure}[htbp]
 \centering
 \includegraphics[width=0.72\linewidth]{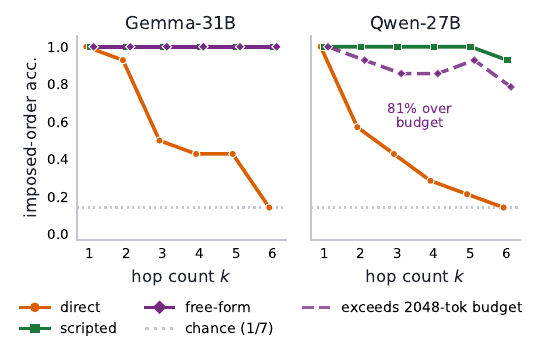}
 \caption{Imposed-order accuracy by hop count under three prompt regimes (days;
 $1/7$ chance dotted): direct, scripted reasoning, and free-form. Qwen free-form runs
 past the 2048-token budget on 81\% of generations; its scripted reasoning reaches
 $0.99$.}
 \label{fig:regimes}
\end{figure}
\Cref{fig:regimes} and \Cref{tab:regimes} report imposed-order accuracy under the three
regimes, with the fraction of generations reaching the 2048-token cap. Across both hero
models, one-shot \emph{direct} reading degrades with hop count while \emph{scripted}
reasoning recovers ceiling (Gemma-31B 1.00, Qwen-27B 0.99); \emph{free-form} recovers
ceiling in Gemma-31B (1.00).

The one asymmetry, Qwen's free-form regime (0.89), is neither a budget nor an extraction
artifact. Gemma stays within the 2048-token budget on 100\% of generations, and Qwen does
too under \emph{scripted} reasoning (solving 0.99), so the map is traversable and the
budget is not the bottleneck; and three answer extractors (last entity, last bolded item,
explicit-conclusion rule) all return 0.89. Instead, all nine free-form misses (of 84 generations) are
truncated with no conclusion---cut mid-deliberation, in every case looping on whether the
order is $0$- or $1$-indexed. Qwen's shortfall is thus a failure of free-form reasoning to
terminate and resolve that ambiguity within budget, not a failure to traverse the imposed
map (which scripted confirms at 0.99).

\begin{table}[htbp]
\centering\small
\caption{Imposed-order accuracy by prompt regime (days, list, bare answers,
$\textrm{max\_new}=2048$). ``over budget'' is the fraction of generations that
reach the 2048-token cap. Gemma terminates within budget and solves both
reasoning regimes; Qwen's free-form reasoning over-runs the budget but its scripted
reasoning does not.}
\label{tab:regimes}
\begin{tabular}{@{}l ccc c@{}}
\toprule
& \multicolumn{3}{c}{imposed accuracy} & \\
\cmidrule(lr){2-4}
model & direct & scripted & free-form & over budget\\
\midrule
Gemma-31B & 0.57 & 1.00 & 1.00 & 0\%\\
Qwen-27B & 0.44 & 0.99 & 0.89 & 81\%\\
\bottomrule
\end{tabular}
\end{table}

\section{Activation-patching details}
\label{app:patching}
\paragraph{Where the map forms: an identity$\rightarrow$binding stage.}
\label{app:localize}
Entity-substitution patching also localizes \emph{where} the map forms, the
one thing the causal probe adds over prompt-editing. A wrong-slot control (inject the
operand one token off the entity) exposes a boundary: at early layers ($\lesssim$0.2
depth) the wrong-slot injection \emph{still} flips the answer (the operand is
token-local \emph{identity}, not yet tied to a map position), while by $\approx$0.45
depth only the entity slot is causal (identity now \emph{bound} into the map), after
which the answer commits by $\approx$0.7 depth. Sampling the control at two live layers
brackets this identity$\rightarrow$binding transition rather than pinpointing it (a full
sweep and a second model/concept are pending; \Cref{fig:crossover}).

\paragraph{Source activation and injection.} For each entity we cache, in a single
forward pass over the conflict-context prompt, the residual-stream activation at that
entity's token position(s) at every swept layer; this patch activation therefore
already carries the current context. To patch $E_{\text{orig}}\!\leftarrow\!E_{\text{patch}}$
we overwrite, via a forward hook at one layer, the residual at the orig entity's
position(s) with the patch entity's, then generate and check whether the answer
becomes the patch entity's successor \emph{under the current context's order}.

\paragraph{Multi-token entities.} Residuals are overwritten position-wise across
the entity's sub-tokens, right-aligned to the trailing positions (the last
$\min(m_{\text{orig}},m_{\text{patch}})$ sub-tokens when patch and orig differ in length), so entities
of unequal sub-token length are matched at their final positions; the reported
behavioral and topology results use the same last-token-pre-generation readout
throughout.

\paragraph{Normalization and the norm confound.} We inject the
\emph{raw} patch residual, with no rescaling. The concern that a flip could be
driven by activation \emph{norm} rather than content is controlled directly by
the random-vector control: a Gaussian vector rescaled to the per-row norm of the
real patch, injected at the same slot, flips the answer only at chance ($0.14$),
whereas the real patch flips it at $1.00$. Norm alone therefore does not produce
the effect.

\paragraph{Chain-of-thought patch (Qwen-27B).} When the model answers by reasoning
step-by-step, the crossover is read at the early-to-mid layers where the patch takes
effect: patching drives the final answer to the imposed-order successor ($1.00$) and
not the pretrained one ($\le0.14$). Deeper patches make this verbose model loop
without committing, so we read at the live layers; the committed answer is scored by
an independent LLM judge rather than a string match.

\paragraph{Step-count sweep.} A companion patch targets the step-count token rather
than the entity: for the query ``advance $k$ steps from $E$'', we overwrite the
$k$-digit residual $k_{\text{orig}}\!\leftarrow\!k_{\text{patch}}$ and check whether the answer
becomes the $k_{\text{patch}}$-hop successor, restricted to pairs correct at both counts.
Run across the ladder (days, $k$ up to 6, 2 scrambles), the step count is causally
used, cleanly in the larger models (patch success $1.00$ for Gemma-12B/31B, $0.96$ for
Qwen-27B, $0.84$ for Qwen-9B) and partially in the smallest (Gemma-E2B $0.62$, E4B
$0.46$, with the query token leaking). So both operands of the traversal (the
entity and the step count) are read from the imposed map.

\paragraph{Mechanism class.} We patch the full residual stream, not attention or
MLP outputs separately, so we localize map formation by \emph{depth}
(\Cref{app:localize}) rather than by component; attention-vs-MLP attribution is
left to future work.

\paragraph{Swept layers.} The sweep is at depth fractions
$\{0.1,0.2,0.3,0.45,0.6,0.7,0.75,0.85,0.95\}\times n_{\text{layers}}$, the $0.75$ entry being the map layer. \Cref{tab:patchlayers} gives the absolute indices per model.
These fractions are fixed a priori (the same grid for every model) and we
report the patch and control success across the whole sweep
(\Cref{fig:crossover}) rather than at a single layer chosen on the outcome, so the
reported dissociation is not selected post hoc.

\begin{table}[htbp]
\centering\small
\caption{Absolute layer indices swept by the activation-patching arm, per model
(the depth fractions of \Cref{app:localize} resolved to integer layers).}
\label{tab:patchlayers}
\begin{tabular}{@{}l c l@{}}
\toprule
model & $n_{\text{layers}}$ & swept layer indices\\
\midrule
Gemma-E2B & 35 & 4, 7, 10, 16, 21, 24, 26, 30, 33\\
Gemma-E4B & 42 & 4, 8, 13, 19, 25, 29, 32, 36, 40\\
Gemma-12B & 48 & 5, 10, 14, 22, 29, 34, 36, 41, 46\\
Gemma-31B & 60 & 6, 12, 18, 27, 36, 42, 45, 51, 57\\
Qwen-27B & 64 & 6, 13, 19, 29, 38, 45, 48, 54, 61\\
\bottomrule
\end{tabular}
\end{table}

\paragraph{The crossover across scale.} \Cref{tab:causalladder} reports the two arms
of the double dissociation across the ladder (2 scrambles $\times$ 42 pairs each). We
separate three things the flagship crossover (\Cref{sec:causal-use}) conflates. \emph{Imp}
is the imposed-context arm (fraction of patched answers equal to the patch's
imposed-order successor); \emph{Nat} is the natural-context arm (its pretrained-order
complement); each is read at the layer where that arm is strongest. \emph{Clean
locus} marks whether a \emph{deep} layer exists that is simultaneously live (imposed
arm $\ge0.9$) and passes the wrong-slot position control ($<0.2$), i.e.\ a genuine
read of the entity rather than a trivial vector injection; where it exists we give its
depth fraction. At 2--9B the patch is effective only at early layers and commits away
by mid-depth, family-dependently: small Qwen carries both arms at an early, leaky layer
(a rough, shallow dissociation), while small Gemma carries only the imposed arm (its
natural-context patch never takes, at any swept layer). The \emph{clean} double
dissociation (both arms high at a clean, deep locus) appears only at Gemma-12B/31B and
Qwen-27B; within Gemma the natural arm comes online with scale. The
mid-ladder entries sit near the $0.9/0.2$ thresholds, so the precise onset scale
should be read as an interval, not a sharp point.

\begin{table}[htbp]
\centering\small
\caption{\textbf{Causal patch success across the scale ladder} (direct prompt, 2
scrambles $\times$ 42 pairs). \emph{Imp}/\emph{Nat}: imposed- and natural-context
arms (own-map success), each at its strongest layer. \emph{Clean locus}: depth of the
deepest layer that is both live ($\text{Imp}\ge0.9$) and passes the wrong-slot control
($<0.2$); ``---'' means no clean deep locus (the flip is one-sided and/or leaky),
and ``n/r'' marks a model$\times$concept not run (Gemma-12B was run on days only).
Chance is $1/7=0.14$ (days), $1/12=0.08$ (months). Llama-8B builds no usable map and
is omitted.}
\label{tab:causalladder}
\begin{tabular}{@{}l rrc rrc@{}}
\toprule
& \multicolumn{3}{c}{days} & \multicolumn{3}{c}{months}\\
\cmidrule(lr){2-4}\cmidrule(lr){5-7}
model & Imp & Nat & clean locus & Imp & Nat & clean locus\\
\midrule
Gemma-E2B & $0.90$ & $0.00$ & --- & $0.78$ & $0.00$ & ---\\
Gemma-E4B & $0.89$ & $0.48$ & --- & $0.95$ & $0.67$ & ---\\
Gemma-12B & $1.00$ & $1.00$ & $0.46$ & n/r & n/r & n/r\\
Gemma-31B & $1.00$ & $1.00$ & $0.45$ & $1.00$ & $0.98$ & $0.45$\\
\addlinespace
Qwen-4B   & $0.73$ & $1.00$ & --- & $0.74$ & $1.00$ & ---\\
Qwen-9B   & $0.89$ & $1.00$ & --- & $0.88$ & $1.00$ & ---\\
Qwen-27B  & $1.00$ & $1.00$ & $0.45$ & $1.00$ & $0.98$ & $0.45$\\
\bottomrule
\end{tabular}
\end{table}

\section{Wrap relation vs.\ endpoint co-occurrence}
\label{app:cooccur}
\paragraph{Measuring cyclic closure.} We gauge how ring-like a geometry is with two
graded indices rather than a binary line/cycle verdict: \emph{dRSA} $=$ RSA(cyclic)
$-$ RSA(linear), signed (higher $=$ more circular, negative $=$ less); and a wrap-pair
\emph{closure ratio} adapting the horseshoe diagnostic of \citet{diaconis2008}, the
last$\leftrightarrow$first distance relative to typical adjacent pairs ($\approx$1 when
tightly closed, $\gg$1 when open). Both report position on an open-to-closed continuum;
persistent homology \citep{carlsson2009} and circular coordinates \citep{desilva2011}
could also serve, and we add a plotted geometry.

\begin{figure}[htbp]
 \centering
 \includegraphics[width=\linewidth]{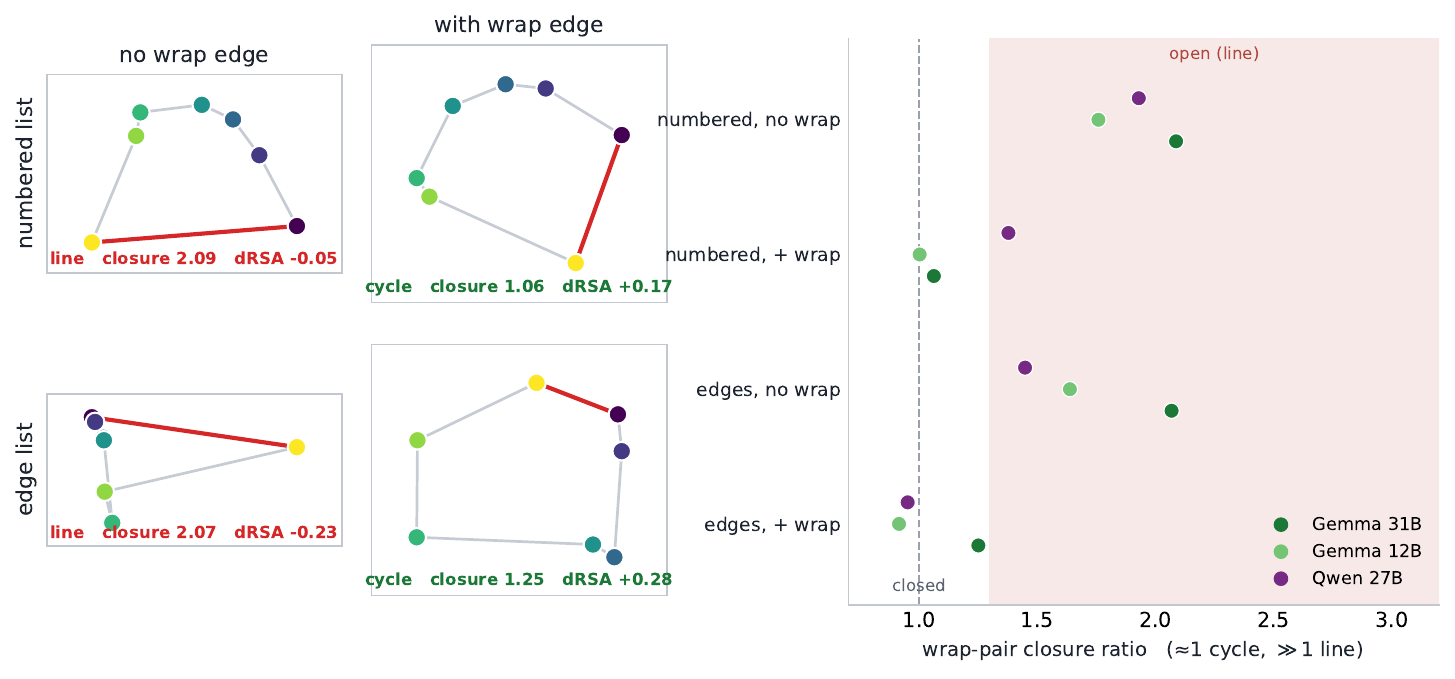}
 \caption{\emph{Left:} form $\times$ wrap 2$\times$2 for Gemma-31B (PC1--PC2 of
 entity centroids, colored by imposed position, wrap edge red); per cell, closure
 $\approx$1 / dRSA $\ge 0$ when the wrap is stated, closure $\gg$1 / dRSA $<0$ when
 open. \emph{Right:} the wrap-pair closure ratio for three models, open vs.\
 wrap-stated.}
 \label{fig:topology}
\end{figure}

\paragraph{The wrap relation strengthens closure.} Holding tokens and task fixed, stating
the \emph{wrap} makes the geometry more ring-like \emph{by degree}: it raises dRSA (from
$<0$ toward $\ge0$) and lowers the closure ratio (from $\gg$1 toward $\approx$1), and
removing it leaves the geometry less circular. The form~$\times$~wrap 2$\times$2
(\Cref{fig:topology}) shows the driver is the stated relation, not the surface format; the
direction is consistent across five concepts and holds on arbitrary tokens with no
pretrained ring to inherit (\Cref{tab:arb}), though its magnitude is capability-dependent
(Gemma-31B 5/5 concepts; Qwen-27B 4/5, clock marginal).

\paragraph{Relation, not co-occurrence.} Because the wrap edge names the two endpoints
together and the closure metric reads exactly that pair, the effect could in principle be
local token co-occurrence. \Cref{tab:cooccur} rules this out with two extra cells:
\emph{co-mention} (endpoints named, no relation) leaves the geometry open (closure
$\approx$ the bare list), so co-occurrence alone does not close the loop; \emph{wrap-by-position}
(the wrap asserted via position indices, endpoint names never co-listed) still pulls toward
a cycle in the capable model, and the adjacent phrasing closes it fully. So the relation sets
the topology and adjacency only adds strength; Qwen-27B is the weaker case, where even the
adjacent wrap is marginal.

\begin{table}[htbp]
\centering\small
\caption{Wrap relation vs.\ endpoint co-occurrence (days, 3 scrambles). closure
$\approx$1 is fully closed (ring), $\gg$1 open. Co-mention (endpoints named, no
relation) stays open; wrap-by-position (relation, names not co-listed) closes
partially; the adjacent wrap closes it fully.}
\label{tab:cooccur}
\begin{tabular}{@{}l cc cc@{}}
\toprule
& \multicolumn{2}{c}{Gemma-31B} & \multicolumn{2}{c}{Qwen-27B}\\
\cmidrule(lr){2-3}\cmidrule(lr){4-5}
condition & closure & dRSA & closure & dRSA\\
\midrule
list (no wrap) & 2.09 & $-0.06$ & 1.96 & $-0.29$\\
co-mention (no relation) & 2.19 & $-0.10$ & 1.84 & $-0.25$\\
wrap by position & 1.47 & $+0.06$ & 1.85 & $-0.30$\\
adjacent wrap & 1.09 & $+0.16$ & 1.36 & $-0.08$\\
\bottomrule
\end{tabular}
\end{table}

\section{Additional redraw evidence}
\label{app:redraw}
\begin{table}[htbp]
\centering\small
\caption{Imposed-order RSA on days (Gemma-31B, Qwen-27B). \emph{Top:} under
non-structural probes (bare mention, story), imposed$-$natural RSA. \emph{Bottom:}
imposed-order RSA by hop count, with the $k{=}1$-vs-full-$k$ RDM agreement.}
\label{tab:robust}
\begin{tabular}{@{}lcc@{}}
\toprule
 & Gemma-31B & Qwen-27B\\
\midrule
\multicolumn{3}{@{}l}{\emph{non-structural probe: imposed$-$natural RSA}}\\
\quad neutral mention & $+0.52$ & $+0.41$\\
\quad story           & $+0.46$ & $+0.41$\\
\midrule
\multicolumn{3}{@{}l}{\emph{imposed-order RSA by hop count}}\\
\quad $k{=}1$ only    & $+0.67$ & $+0.64$\\
\quad $k{=}6$ only    & $+0.71$ & $+0.54$\\
\quad full $k$        & $+0.81$ & $+0.56$\\
\quad $k{=}1$ vs.\ full-$k$ RDM ($\rho$) & $0.74$ & $0.93$\\
\bottomrule
\end{tabular}
\end{table}
\begin{figure}[htbp]
  \centering
  \includegraphics[width=\linewidth]{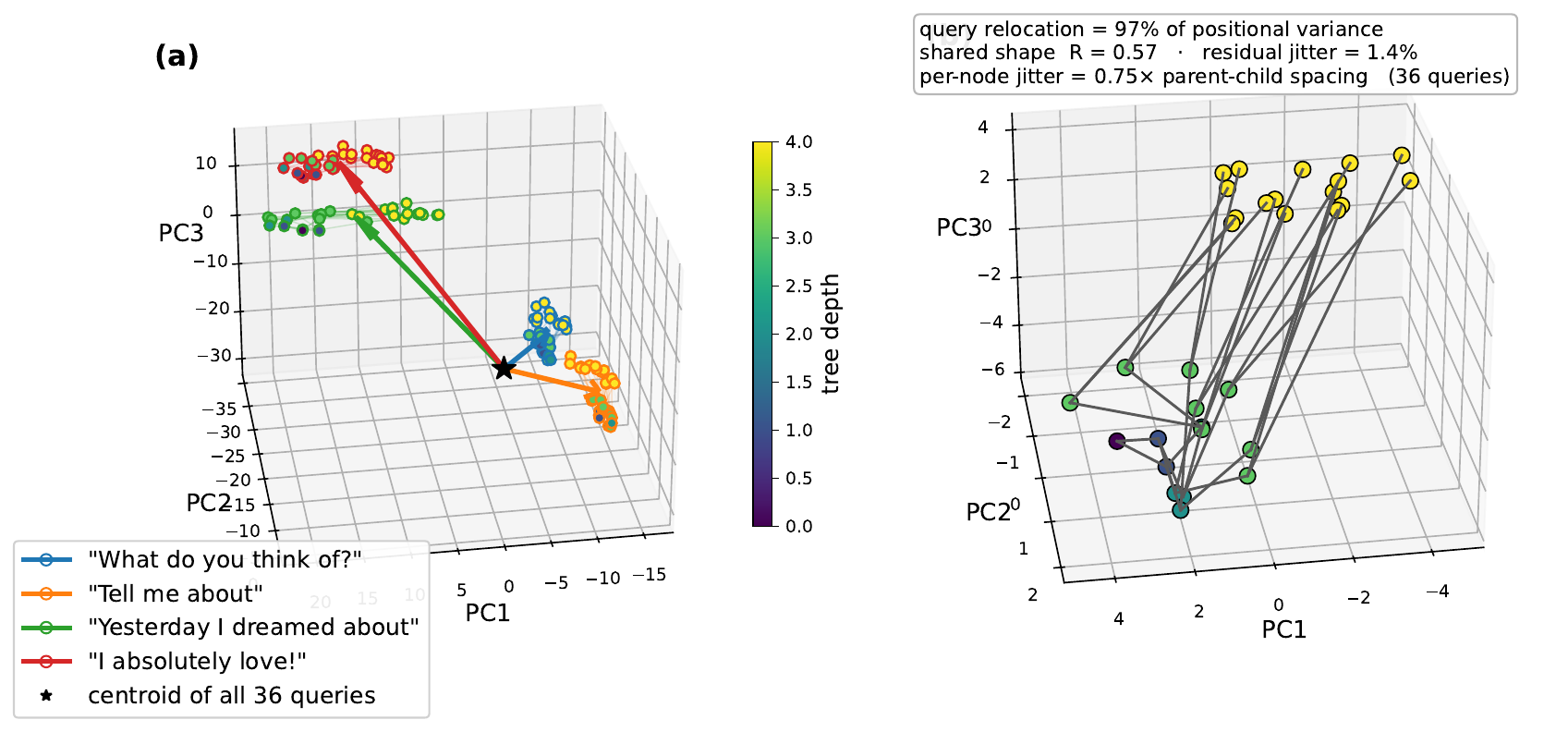}
  \caption{Imposed depth-4 tree across 36 query types (Gemma-31B). \emph{(a)} Four
    query types place the tree at different locations (arrows $=$ query vectors from
    the 36-query centroid; color $=$ depth). \emph{(b)} The per-node centroids, recentered to a common origin ($\rho=0.70$ to the consensus; per-node jitter
    $0.75\times$ parent--child spacing); the query accounts for 97\% of positional
    variance.}
  \label{fig:queryreloc}
\end{figure}
\begin{table}[htbp]
\centering\small
\caption{Per-model imposed-order vs.\ residual natural-order RSA (direct prompt,
means over 10 scrambles): the evidence behind the dominance (\Cref{sec:redraw}) and
within-family scale claims. \Cref{fig:dominance} plots the days columns.}
\label{tab:rsa}
\begin{tabular}{@{}l rr rr@{}}
\toprule
& \multicolumn{2}{c}{days} & \multicolumn{2}{c}{months}\\
\cmidrule(lr){2-3}\cmidrule(lr){4-5}
model & imposed & natural & imposed & natural\\
\midrule
Gemma-31B & $+0.87$ & $-0.03$ & $+0.81$ & $+0.05$\\
Gemma-12B & $+0.81$ & $-0.03$ & $+0.54$ & $+0.14$\\
Gemma-E4B & $+0.72$ & $+0.06$ & $+0.58$ & $+0.31$\\
Gemma-E2B & $+0.60$ & $-0.06$ & $+0.59$ & $+0.23$\\
\addlinespace
Qwen-27B  & $+0.67$ & $-0.01$ & $+0.71$ & $+0.12$\\
Qwen-9B   & $+0.10$ & $+0.44$ & $+0.60$ & $+0.19$\\
Qwen-4B   & $+0.31$ & $+0.24$ & $+0.46$ & $+0.38$\\
\bottomrule
\end{tabular}
\end{table}
The per-model imposed-order vs.\ residual natural-order RSA is given in
\Cref{tab:rsa}. Beyond that summary the redraw is supported by per-concept,
per-scramble \emph{ring decks} (the imposed-order ring drawn for each scramble),
and by the two confound controls (the crossover and the anisotropy-matched
null) whose full procedure and numbers are given in \Cref{app:controls}.
Per-scramble confidence intervals for all headline numbers accompany the release.

\paragraph{RSA is not a projection artifact.} RSA is computed in the full
residual space; the 2D and 3D plots are illustration. \Cref{tab:proj} shows
that for the displayed geometries 2D, 3D, and full-dimensional RSA agree, so the
low-dimensional views are faithful: in particular, the third PC does not recover
a ring on the weak models (Qwen-9B and Qwen-4B, RSA $\approx$0.25 in 2D, 3D, and
full alike; Gemma reorganizes even at E2B, which is why it sits higher). A clean,
cycle-like geometry appears only in the most capable models, and those also carry
the most \emph{concentrated} variance (the top three PCs explain $\sim$0.85),
whereas the weak models are diffuse ($\sim$0.65) with no ring in any projection.

\begin{table}[htbp]
\centering\small
\caption{Imposed-order RSA at increasing projection dimension (days, scramble
$s_0$), with the variance captured by the top three PCs. 2D/3D/full agree; the
clean cycle and concentrated variance coincide at the largest models.}
\label{tab:proj}
\begin{tabular}{@{}l rrr r@{}}
\toprule
model & RSA 2D & RSA 3D & RSA full & top-3 var.\\
\midrule
Gemma-31B & $+0.81$ & $+0.85$ & $+0.82$ & 0.88\\
Gemma-E2B & $+0.55$ & $+0.64$ & $+0.56$ & 0.71\\
Qwen-27B  & $+0.78$ & $+0.71$ & $+0.65$ & 0.83\\
Qwen-9B   & $+0.28$ & $+0.23$ & $+0.26$ & 0.71\\
Qwen-4B   & $+0.32$ & $+0.31$ & $+0.24$ & 0.63\\
\bottomrule
\end{tabular}
\end{table}

\paragraph{Topology on arbitrary tokens.} \Cref{tab:arb} gives the 2$\times$2 on
arbitrary, structure-free tokens (no pretrained cyclic order), the control behind
the claim in \Cref{sec:topology} that the wrap statement, not any pretrained
structure, drives the closure. Adding the wrap closes the loop (dRSA $\ge 0$,
closure $\approx$1) and removing it leaves the geometry open, less circular (dRSA
$<0$, closure $\gg$1), in both surface forms.

\begin{table}[htbp]
\centering\small
\caption{Topology on arbitrary, structure-free tokens (Gemma-31B): dRSA and
wrap-pair closure for the four specifications. The wrap relation, not the surface
form or any pretrained structure, sets the topology.}
\label{tab:arb}
\begin{tabular}{@{}l rr@{}}
\toprule
specification & dRSA & closure\\
\midrule
numbered list, no wrap   & $-0.12$ & 2.7\\
numbered list, $+$wrap   & $+0.16$ & 1.06\\
edges, no wrap           & $-0.27$ & 3.3\\
edges, $+$wrap           & $+0.31$ & 1.1\\
\bottomrule
\end{tabular}
\end{table}

\section{Confound controls in detail}
\label{app:controls}
This appendix gives the full procedure and numbers for the relabeling and anisotropy
confounds summarized in \Cref{sec:redraw} (the third, serial position, is handled by the
tree and wrap evidence of \Cref{sec:topology,app:cooccur}); numbers are Gemma-31B on days
unless noted, from the same conflict-state residuals used throughout. Both controls address
the residual stream's strong \emph{anisotropy}: the entity centroids share a dominant
direction, so raw cosines are high and barely distinguish the orderings (imposed-adjacent
pairs average 0.986 vs.\ 0.979 natural-adjacent, a 0.007 gap). The reported structure lives
not there but in the \emph{mean-centered} RDM, which removes the shared direction before
measuring proximity; every RSA in the paper uses it. The controls verify the resulting
signal is neither a renaming of a fixed geometry (Control~1) nor an artifact of the cone
(Control~2).

\paragraph{Control 1: relabeling (does the geometry actually move?).} \emph{Worry:} a high
imposed-order RSA might only reflect renaming which point we call ``position~3,'' with a
fixed set of centroids scoring equally against any declared order. \emph{Test:} probe two
identical-token contexts (natural, scrambled) and score each context's centroids against
both templates; a genuine redraw predicts a crossover, a fixed-but-relabeled geometry
predicts none. \emph{Result:} the crossover is sharp---scrambled context imposed-RSA
$+0.84$/natural $-0.04$, natural context imposed $-0.10$/natural $+0.87$---so the same
tokens give opposite geometries. Ranked by RSA to the scrambled-context geometry against all $7!=5040$ orderings, the imposed order ties for first (15 orderings share the maximum RSA, top 0.3\%): among all orderings the geometry fits the imposed one best.

\paragraph{Control 2: anisotropy (is the RSA an artifact of the cone?).} \emph{Worry:} in a
narrow cone even an unstructured arrangement can give a positive RSA, so we need a null with
the data's cone geometry but no imposed structure. \emph{Test (label-shuffle null):} shuffle
which entity label sits on which real centroid vector, mean-center, and recompute the
imposed-order RSA, $2{,}000$ times per scramble across ten orders; this preserves the cone,
norms, and pairwise angles while destroying the geometry--order correspondence, and we
report the empirical tail probability rather than a Gaussian $z$. \emph{Result:} the observed
RSA clears this null on every concept at both flagships---all ten scrambles significant for
days, notes, months, zodiac, and clock, natural-order retention near zero ($-0.05$ to
$+0.12$)---with the margin growing in $N$: the twelve-point concepts clear at $p<0.001$ (no
draw of the $2{,}000$ reaches the observed RSA), the seven-point concepts at $p\le0.005$
(Gemma-31B) and $p\le0.008$ (Qwen-27B).

\emph{Caveat (the smallest concept).} At $N=7$ the null has a small floor, and it is
\emph{structural}, not noise. A cyclic template is invariant to rotation and
reflection, so the imposed $7$-cycle is matched by its $2\times7=14$ symmetry images
(the same ring, relabeled by where it starts and which way it runs, all giving an
identical distance template); and because the cyclic distance takes only three values
at $N=7$ ($1,2,3$), a few near-orderings tie as well. These---not spurious fits---are
the $\sim$$15$ of $2{,}000$ draws that reach the observed RSA, the same sense in which
the imposed order ties for first among all $5040$ orderings (Control~1). Every days
scramble still clears at $p\le0.008$, and at $N=12$ the $24$-ordering orbit is a
vanishing fraction of $12!$ ($5\times10^{-8}$), effectively never sampled, so
$p\approx0$. The null holds on every concept, its small-$N$ floor set by cycle
symmetry rather than weak geometry.

\section{RSA carries structure: convergent validity}
\label{app:convergent}
Because RSA-against-a-template is not the standard tool in this subfield (which
relies on PCA inspection and probes, \Cref{sec:related}), we verify that our
imposed-order RSA tracks \emph{the same structure} that the field's own measures
recover, by cross-checking it against four independent quantities. They agree.

\begin{itemize}[leftmargin=*]
 \item \textbf{Field-standard methods.} The circular-manifold result our work
 builds on was established via sparse autoencoders and PCA inspection
 \citep{engels2024}; our RSA recovers the \emph{same} imposed ring on the same
 concepts. RSA quantifies, with a significance test, what those methods show by
 inspection.
 \item \textbf{Ring planarity (PCA).} Across the eight models, imposed RSA
 rank-tracks the cleanliness of the PCA ring (fewer self-crossings) at
 Spearman $\rho=0.93$: a high RSA is a clean, low-crossing circle, a low RSA a
 tangled one.
 \item \textbf{Behavior.} Imposed RSA rank-tracks one-shot adjacency accuracy at
 $\rho=0.83$ across the eight models, the geometry tracks what the model can
 \emph{do} with the map, not just how it looks.
 \item \textbf{Persistent homology.} On the imposed cycle (arbitrary tokens) the
 centroid cloud carries a dominant persistent $H_1$ bar at $8$--$11\times$ the
 top-bar persistence of random points of matched $N$ and dimension (Gemma-31B
 $11.5\times$, Gemma-E2B $8.4\times$), whereas the imposed tree carries no comparable
 loop (Gemma-31B $2.0\times$, at the random-baseline floor; Qwen-27B $4.7\times$, a
 weaker but still-present separation), a rotation- and relabeling-invariant
 confirmation that the cycle and the tree are different topology \emph{types}
 \citep{trsa2024,carlsson2009}. We report the persistence \emph{ratio} rather than a
 thresholded Betti count, and treat it as a confirmatory diagnostic, not a primary
 test: at $N{=}7$--$12$ the barcode is sparse, so dRSA and the closure ratio remain
 the primary topology discriminators. This is, to our knowledge, the first use of
 persistent homology for \emph{concept-level} topology detection in language
 models (it has been applied to bulk representations elsewhere).
\end{itemize}

\noindent Four independent measures (a fitted probe, PCA planarity, behavior,
and a topological invariant) converge on the structure RSA reports, which is
strong evidence the metric carries genuine relational information rather than a
high-dimensional artifact. The two rank correlations here ($\rho=0.93$ planarity,
$\rho=0.83$ behavior) are computed across eight models spanning families and
scale, so they are cross-family, rank-level agreement (consistent with, but not
independent of, the scale trend) rather than a within-model prediction; we use
them as convergent validity, not as an effect size. The anisotropy control and the rationale for
mean-centering over whitening or CKA are in \Cref{app:controls};
\citet{ethayarajh2019,meancentring2023} establish that residual representations
are anisotropic and that mean-centering is the appropriate lightweight correction.

\section{Topology range: not just cycles}
\label{app:grid}
The formation-and-override result is not specific to 1-D cyclic concepts; it
extends to 2-D grids, reported with the same full-dimensional RSA, permutation
null, and imposed-vs-natural dominance used for the cyclic concepts.

\paragraph{An imposed 2-D layout overrides a native 2-D prior.} On a 3$\times$3
block of QWERTY keys (Q W E / A S D / Z X C) the model's native keyboard geometry
is a weak 2-D grid, keyboard-distance RSA $+0.53$ at Qwen-27B, beating the 1-D
nulls (reading-order $+0.20$, alphabetical $+0.31$; $p{=}0.007$), and weak at
Gemma-31B. (These are single-letter keys, so the signal is spatial, not a shared
sub-token artifact.) Imposing a redefined layout (shuffled ``right of''/``below''
relations) overrides it: imposed-layout vs native-QWERTY RSA $+0.78$ vs $+0.09$ at
Qwen-27B (dominance $+0.69$, 6/6 scrambles significant against the null) and $+0.42$
vs $+0.02$ at Gemma-31B (dominance $+0.40$, 5/6), the 2-D analog of the
imposed-order dominance on cyclic concepts.

\paragraph{Formation on arbitrary tokens.} Imposing a 3$\times$3 grid on
meaning-free tokens reorganizes their geometry toward it: grid-distance RSA exceeds
the 1-D line template (Gemma-31B $+0.43$, beating the line on 6/6 scrambles), so the
context forms a 2-D structure with no prior to inherit.

\paragraph{Scope.} The model's native 2-D grids (keyboard, chess) are weaker and
more model-dependent than its native cyclic rings, so the cyclic concepts carry the
main results; the 2-D evidence shows the formation-and-override mechanism is not
tied to a single topology.

\section{Shapes on a fixed concept (days): cross-family and method}
\label{app:shapes}
On the seven weekdays we impose a cycle, a line, or a binary tree and read the
geometry at layer $0.75\,n$, last-token pre-generation, over 12 neutral/diverse
queries and 10 scrambles (\Cref{sec:topology}, \Cref{fig:shapes}). \emph{Effective
dimension} is the participation ratio $(\sum_i\lambda_i)^2/\sum_i\lambda_i^2$ of the
entity-centroid PCA eigenvalues, an ordering-free line-versus-ring discriminator
($\approx$1 for a line, $\approx$2 and above for a ring) that we prefer to dRSA
(weak inside a ring). The ring and the tree are categorical in both flagships; the
line collapses to $\approx$1.5-D in Gemma but only partially in Qwen ($\approx$2.7-D),
so the clean line is capability-dependent.

\begin{table}[htbp]
\centering\small
\caption{Shapes imposed on the seven weekdays (means over 10 scrambles; NAT is the
un-imposed baseline, read at the day token). ``key RSA'' is cyclic-RSA for
NAT/cycle/line and depth-RSA for tree; effective dimension is the participation
ratio.}
\label{tab:shapes}
\begin{tabular}{@{}l cc cc@{}}
\toprule
& \multicolumn{2}{c}{Gemma-31B} & \multicolumn{2}{c}{Qwen-27B}\\
\cmidrule(lr){2-3}\cmidrule(lr){4-5}
imposed spec & key RSA & eff-dim & key RSA & eff-dim\\
\midrule
NAT (ring)   & $+0.59$ & 4.8 & $+0.77$ & 5.0\\
cycle (ring) & $+0.72$ & 4.0 & $+0.71$ & 3.8\\
line         & $+0.39$ & \textbf{1.5} & $+0.65$ & 2.7\\
tree (depth) & $+0.67$ & 1.8 & $+0.68$ & 2.1\\
\bottomrule
\end{tabular}
\end{table}


\end{document}